**Title:** A Machine Learning model of the combination of normalized SD1 and SD2 indexes from 24h-Heart Rate Variability as a predictor of myocardial infarction

Antonio Carlos Silva-Filho[1,2,4], Sara Raquel Dutra-Macêdo[1,3], Adeilson Serra Mendes Vieira[1,3], Cristiano Mostarda[1,3]

1 Laboratory of Cardiovascular Adaptations to Exercise, Universidade Federal do Maranhão, São Luís, Brazil

2 Doutorado da Rede Nordeste em Biotecnologia – RENORBIO, Universidade Federal do Maranhão, São Luís, Brazil

3 Programa de Pós-graduação em Saúde do Adulto, Universidade Federal do Maranhão, São Luís, Brazil

4 Faculdade Uninassau, São Luís, Brazil

Abstract

**Aim:** to evaluate the ability of the nonlinear 24-HRV as a predictor of MI using Machine Learning

**Methods:** The sample was composed of 218 patients divided into two groups (Healthy, n=128; MI n=90). The sample dataset is part of the Telemetric and Holter Electrocardiogram Warehouse (THEW) database, from the University of Rochester Medical Center. We used the most common ML algorithms for accuracy comparison with a setting of 10-fold cross-validation (briefly, Linear Regression**,** Linear Discriminant Analysis, *k*-Nearest Neighbor, Random Forest, Supporting Vector Machine, Naïve Bayes, C 5.0 and Stochastic Gradient Boosting). **Results:** The main findings of this study show that the combination of SD1nu + SD2nu has greater predictive power for MI in comparison to other HRV indexes. **Conclusion:** The ML model using nonlinear HRV indexes showed to be more effective than the linear domain, evidenced through the application of ML, represented by a good precision of the Stochastic Gradient Boosting model.

**Introduction**

Myocardial infarction (MI) is the most common type of cardiovascular disease (CVD) and one of the leading causes of death in the world After MI (1), local changes are observed in cardiac tissue, such as ventricular remodeling characterized by left ventricular dilation, changes in ventricular wall structure, increased muscle mass, and decreased cardiac function (2). Also, the autonomic nervous system imbalance is observed after MI systemic changes, with a marked increase in sympathetic modulation and reduction of vagal (parasympathetic) modulation (3)

The risk assessment of patients with MI is fundamental to identify and choose optimal therapeutic strategies to improve the outcome of each patient. The analysis of heart rate variability (HRV) has become a popular diagnostic tool in clinical practice to evaluate the modulation of the autonomic nervous system. HRV is the consecutive variation of the RR intervals of the electrocardiogram in a specific duration that correlates with the sympathetic, parasympathetic and, thereafter, sympathovagal balance (4,5) HRV analysis has been used for many years to measure autonomic modulation because of its simplicity, accuracy and non-invasive nature (6) However, the use of HRV for the diagnosis of acute myocardial infarction has not been evaluated.

Currently, the heart rate variability (HRV) study has become an important noninvasive diagnostic tool in cardiology to evaluate the activities of the ANS because it is a simple, reliable and non-invasive method of monitoring. In fact, over the past 40 years, HRV has been used to diagnose autonomic dysfunction and to quantify the associated risk in a variety of cardiac and noncardiac disorders (7,8). Particularly in patients with MI, HRV has been extensively studied and has been shown to have an important prognostic value (9).

The evaluation of the ANS through nonlinear measurements has been used to clarify the complexity of the data, being much closer to the human biodynamic system, which behaves in a nonlinear way (10,11). Therefore, several studies analyzed HRV through different methods of digital signal processing and statistical methods (12,13)

Although some nonlinear measurements used in the analysis of autonomic modulation (AM) may be adequate using short-term series, they are generally dependent on long-term data series, as was the case with Shannon Entropy (14) and correlation dimension (15). Although there are many methods in the field of chaos (16), other methods may help to understand the complex field of HRV. In this context, chaotic global methods were formulated by Garner e Ling (17), who proposed a robust method of chaos analysis to investigate disease conditions and the evolution of therapeutic interventions.

Machine learning methods, for example, are extremely powerful tools for classifying and predicting binary and/or multiclass heart disease. Many previous studies have attempted to automatically distinguish patients with a particular disease from patients with other diseases or healthy controls from machine learning methods (18). Thus, in this study, we intend to evaluate the ability of the nonlinear 24-HRV as a predictor of MI by comparing the area under the curve, sensitivity, and specificity and Cohen's Kappa.

**Materials and Methods**

*Sample*

The sample was composed of 218 patients divided into two groups (Healthy, n=128; MI n=90). The sample dataset is part of the Telemetric and Holter Electrocardiogram Warehouse (THEW) database, from the University of Rochester Medical Center. After an official contact with the THEW coordinators, we downloaded and analyzed the data using the HRVanalysis software (7). The database E-HOL-03-0160-001 was composed of 93 24h-Holter recordings from subjects with MI. For the Healthy group, we used the E-HOL-03-0202-003 database of 130 24h-Holter recordings of healthy subjects.

*Machine Learning (ML) models*

The R software (19) was used for the ML models analysis, with the help of the packages *caret*, *relaimpo* and *mlbench*. First, the dataset was split into two groups for algorithm validation. The first part was composed of 80% of the data and was used as the asset for training the ML algorithms. The second part (the labeled remaining 20%) was used for the ML algorithm validation and accuracy evaluation. Then, we chose the most common ML algorithms for accuracy comparison with a setting of 10-fold cross-validation (briefly, Linear Regression, Linear Discriminant Analysis, *k*-Nearest Neighbor, Random Forest, Supporting Vector Machine, Naïve Bayes, C 5.0 and Stochastic Gradient Boosting).

*Heart Rate Variability*

An electrocardiograma used to analyze HRV (Micromed Wincardio 600hz, Brasilia, Brazil).Thespectrum resulting from the Fast Fourier Transform model is derived from all the data present from the signal recorded in a minimum of five minutes; the spectrum includes the entire signal variance irrespective of whether the frequency elements of the signals appear as special spectral peaks or as broadband power units.

We used the following HRV indexes: Total variance (ms2), mean of all RR intervals (ms), SDNN (ms, standard deviation of all normal to normal intervals), RMSSD (ms, root mean square of the successive differences of the normal to normal intervals), LF (ms2 and, normalized units, low-frequency spectral density), HF (ms2,and normalized units, high-frequency spectral density), LF/HF (ratio of low and high frequency domains). We also analyzed the Heart Rate Variability (HRV) in the turbulence, deceleration/acceleration capacity, and nonlinear domains.

*Statistical analysis*

The efficacy of the ML models was measured using the Accuracy and Cohen's Kappa, an indicator of the level of false negatives and positives (20). Also, the Area Under de Receiver Operating Characteristic Curve (AUROC) was calculated alongside the sensitivity and specificity of the ML model. We also applied a Two-way ANOVA with a Tukey post-hoc test for the comparison between groups MI and Healthy in Table X. P <0.05 was considered significant.

**Results**

The overall autonomic characteristic of the sample is shown in Table 7, indicating a reduced autonomic modulation in the MI group, when compared to the Healthy group (p<0.05). Among all models and all classification, the one who demonstrated better accuracy and higher kappa values were the Stochastic Gradient Boosting model (Tables 1 to 5). Also, when taking the Stochastic Gradient Boosting alone, the combination of SD1nu + SD2nu performed better as a predictor of MI than Time, Frequency, Nonlinear domains, and Turbulence indexes (Table 6), with values close to 100%. This trend is confirmed even when comparing the combination of SD1nu + SD2nu with SD1nu or SD2nu alone, indicating a great capacity for predicting MI of these variables (Figure 1).

Table 7 shows the 24h, Day and Night HRV analysis of the MI and Healthy groups comparatively. Significant differences were found in the time, turbulence, deceleration/acceleration capacity, and nonlinear domains showing reduced overall cardiac autonomic modulation in the MI group in comparison to the Healthy group (Tukey's p<0.05).

**Discussion**

The present study aimed to evaluate the ability of HRV indexes as predictors of MI using various ML classification algorithms. The main findings of this study show that the combination of SD1nu + SD2nu has greater predictive power for MI in comparison to other HRV indexes. To the best of our knowledge, this is first to evaluate data from 24-h ECG Holter using different models of the ML to verify the predictive potential of myocardial infarction by HRV indexes.

The reduction in HRV after MI is a major risk factor for arrhythmias and death (21). The responses found through HRV indexes may reflect an imbalance between the sympathetic and vagal (parasympathetic) modulation, allowing the prediction of infarcted individuals (21,22). Reduced R-R intervals, reduced time domain indexes (RMSSD, SDNN, and SDANN), increased frequency domain indexes (VLF, LFnu e LF/HF), and the reduction in the complexity of nonlinear methods (SD1, SD2 e ApEn) are correlated with cardiac pathologies such as MI (23–25).

Gyu Lee et al. (2007) (26) found that the combination of linear and nonlinear analysis of HRV demonstrated a greater prediction capacity of coronary artery diseases including MI. In corroboration, a study by Poddar and colleagues (2015) (27) presented better precision and sensitivity in the prediction of MI among other coronary diseases using a combined approach of HRV form that using the indexes alone.

Our data showed that the combination of SD1nu + SD2nu was more effective in predicting the MI than other commonly used HRV domains, such as time and frequency domains (Table 7). The time and frequency domains of HRV have been extensively studied in the last decades and have both strong experimental and clinical data pointing to its usefulness in the clinical practice (28–30). However, recent studies have been pointing out limitations of time and frequency domain methods, such as high susceptibility to noise and stationarity (30,31).To overcome these barriers, nonlinear methods have been studied and are now well understood as representatives of the autonomic modulation (21,32).

Recent studies using ML models with HRV indexes as predictors of MI have shown limitations such as short-term recordings, use of a single ML model (mostly Supporting Vector Machine) and small sample size (18,27). These limitations stop us from making general assumptions of the prediction power of HRV for MI. Our study used a large sample of 24h-Holter recording through different ML models, thus improving the prediction power.

We found that the use of the HRV tool predicts MI, besides it is a non-invasive method, which does not generate discomfort or pain to the patients, being able to be widely used. Concomitant to this, ML models have been widely used in the medical field to assist in the prediction of diseases such as MI since it is a tool with low cost, easy access, and great applicability. In this way, they can aid in the diagnosis and subsequent treatment of cardiac events such as MI.

HRV indexes are predictors of MI, but in the present study, surprisingly, the nonlinear domain indexes SD1nu + SD2nu proved to be more effective than the linear domain, evidenced through the application of ML, represented by a good precision of the Stochastic Gradient Boosting model.

**Table 1.** Combined 24-hour's time domain indexes and its predictions powers according to different models

| Models | Accuracy | Cohen's Kappa | AUROC | Sensitivity | Specificity |
|---|---|---|---|---|---|
| Linear Regression | 0.93 | **0.86** | 0.96 | **0.91** | **0.94** |
| Linear Discriminant Analysis | 0.92 | 0.83 | **0.98** | 0.88 | 0.93 |
| k-Nearest Neighbor | 0.82 | 0.61 | 0.75 | 0.42 | 0.88 |
| Random Forest | 0.93 | 0.86 | 0.98 | 0.90 | 0.92 |
| Supporting Vector Machine | 0.89 | 0.76 | 0.94 | 0.79 | 0.90 |
| Naïve Bayes | 0.84 | 0.66 | 0.90 | 0.81 | 0.87 |
| C 5.0 | 0.92 | 0.83 | **0.98** | **0.91** | **0.94** |
| Stochastic Gradient Boosting | **0.95** | 0.89 | 0.98 | 0.89 | 0.93 |

**Table 2.** Combined 24-hours frequency domain indexes and its predictions powers according to different models

| Models | Accuracy | Cohen's Kappa | AUROC | Sensitivity | Specificity |
|---|---|---|---|---|---|
| Linear Regression | 0.83 | 0.66 | 0.91 | 0.87 | 0.80 |
| Linear Discriminant Analysis | 0.82 | 0.65 | 0.89 | **0.88** | 0.74 |
| k-Nearest Neighbor | 0.78 | 0.55 | 0.91 | 0.73 | 0.85 |
| Random Forest | 0.87 | 0.73 | 0.90 | 0.85 | 0.81 |
| Supporting Vector Machine | 0.80 | 0.59 | 0.91 | 0.81 | 0.83 |
| Naïve Bayes | 0.82 | 0.64 | 0.90 | **0.88** | 0.75 |
| C 5.0 | **0.88** | 0.74 | 0.92 | **0.88** | 0.83 |
| Stochastic Gradient Boosting | **0.88** | 0.74 | 0.94 | 0.86 | **0.86** |

**Table 3.** Combined 24-hours nonlinear domain indexes and its predictions powers according to different models

| Models | Accuracy | Cohen's Kappa | AUROC | Sensitivity | Specificity |
|---|---|---|---|---|---|
| Linear Regression | 0.89 | 0.76 | 0.92 | 0.85 | 0.91 |
| Linear Discriminant Analysis | 0.90 | 0.79 | 0.94 | 0.88 | 0.91 |
| k-Nearest Neighbor | 0.89 | 0.77 | **0.97** | 0.92 | 0.91 |
| Random Forest | 0.94 | 0.87 | **0.97** | 0.93 | 0.93 |
| Supporting Vector Machine | 0.91 | 0.82 | **0.97** | 0.86 | 0.91 |
| Naïve Bayes | 0.85 | 0.69 | 0.95 | 0.85 | 0.85 |
| C 5.0 | 0.94 | 0.87 | 0.95 | 0.89 | 0.93 |
| Stochastic Gradient Boosting | **0.95** | **0.88** | 0.96 | **0.94** | **0.94** |

**Table 4.** Combined 24-hour's turbulence indexes and its predictions powers according to different models

| Models | Accuracy | Cohen's Kappa | AUROC | Sensitivity | Specificity |
|---|---|---|---|---|---|
| Linear Regression | 0.82 | 0.52 | 0.86 | 0.64 | 0.88 |
| Linear Discriminant Analysis | 0.83 | 0.55 | 0.85 | 0.71 | 0.87 |
| *k*-Nearest Neighbor | 0.83 | 0.53 | 0.89 | 0.63 | 0.94 |
| Random Forest | 0.84 | 0.56 | **0.88** | 0.6 | 0.91 |
| Supporting Vector Machine | **0.88** | 0.65 | 0.84 | 0.57 | **0.96** |
| Naïve Bayes | 0.84 | 0.59 | 0.86 | **0.78** | 0.85 |
| C 5.0 | 0.87 | **0.62** | **0.88** | 0.58 | 0.92 |
| Stochastic Gradient Boosting | 0.87 | 0.61 | 0.85 | 0.60 | 0.92 |

**Table 5.** Combined 24-hour's SD1nu + SD2nu and its predictions powers according to different models

| Models | Accuracy | Cohen's Kappa | AUROC | Sensitivity | Specificity |
|---|---|---|---|---|---|
| Linear Regression | 0.95 | 0.89 | **0.98** | 0.94 | 0.95 |
| Linear Discriminant Analysis | **0.96** | **0.91** | 0.97 | **0.99** | 0.92 |
| *k*-Nearest Neighbor | 0.95 | 0.90 | 0.97 | 0.92 | 0.93 |
| Random Forest | 0.94 | 0.88 | 0.97 | 0.98 | 0.95 |
| Supporting Vector Machine | **0.96** | 0.91 | 0.96 | 0.92 | 0.93 |
| Naïve Bayes | 0.94 | 0.89 | **0.98** | 0.95 | 0.94 |
| C 5.0 | **0.96** | **0.91** | 0.97 | 0.95 | 0.93 |
| Stochastic Gradient Boosting | **0.96** | **0.91** | 0.97 | 0.96 | **0.96** |

**Table 6.** Performance of different predictors according to the Stochastic Gradient Boosting model

| Stochastic Gradient Boosting model | Accuracy | Cohen's Kappa | AUROC | Sensitivity | Specificity |
|---|---|---|---|---|---|
| Time domain | 0.95 | 0.89 | 0.98 | 0.89 | 0.93 |
| Frequency domain | 0.88 | 0.74 | 0.94 | 0.86 | 0.86 |
| Nonlinear domain | 0.95 | 0.88 | 0.96 | 0.94 | 0.94 |
| Turbulence indexes | 0.87 | 0.61 | 0.85 | 0.60 | 0.92 |
| SD1nu + SD2nu | **0.96** | **0.91** | **0.97** | **0.96** | **0.96** |

**Figure 1.** Receiver Operator Characteristic curve plot of SD1nu and SD2nu performances

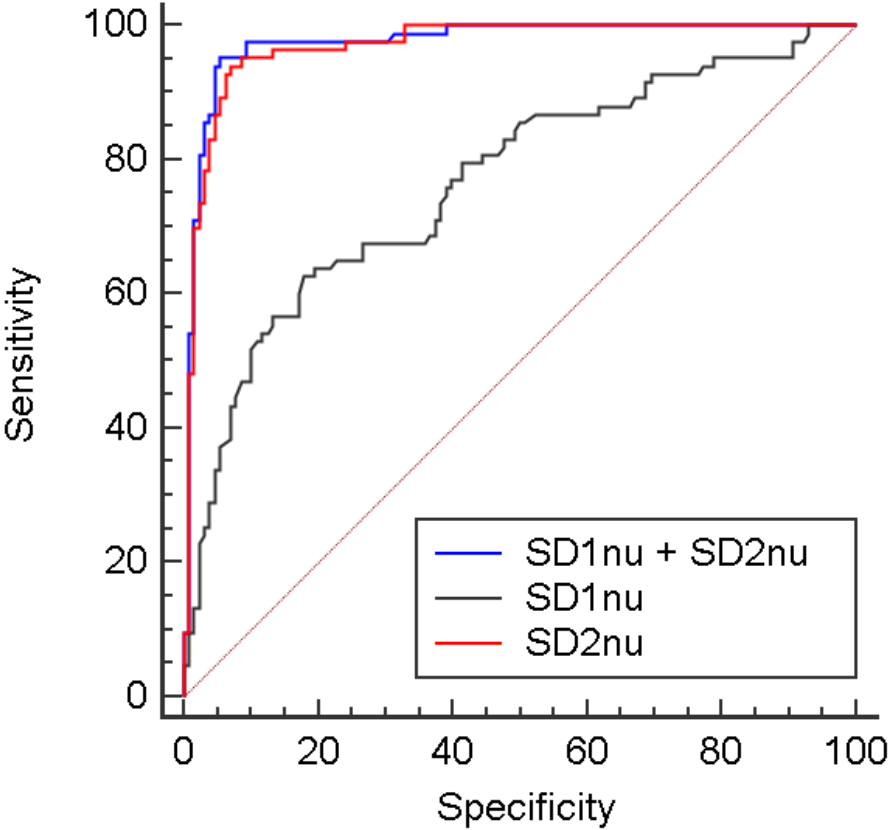

**Table 7.** Overall autonomic modulation of both groups

| | Healthy (n=128) | | | | | | Myocardial infarction (n=90) | | | | | |
|---|---|---|---|---|---|---|---|---|---|---|---|---|
| | 24h | | Day | | Night | | 24h | | Day | | Night | |
| | **Mean** | **SD** | **Mean** | **SD** | **Mean** | **SD** | **Mean** | **SD** | **Mean** | **SD** | **Mean** | **SD** |
| **Time domain** | | | | | | | | | | | | |
| Mean RR (ms) | 790.92 | 101.52 | 733.54* | 93.12 | 934.25*¥ | 143.94 | 870.40† | 127.84 | 854.85† | 198.27 | 898.84 | 211.59 |
| Mean HR (bpm) | 77.12 | 9.98 | 83.11* | 10.56 | 65.75*¥ | 10.23 | 70.43† | 10.43 | 71.73† | 16.64 | 68.33 | 16.11 |
| pcNN20 (%) | 40.06 | 16.25 | 34.62 | 16.25 | 53.67*¥ | 19.26 | 28.05† | 16.85 | 27.09† | 17.78 | 31.05† | 18.55 |
| pcNN30 (%) | 27.47 | 15.20 | 22.39 | 14.27 | 40.30*¥ | 20.56 | 15.76† | 13.56 | 15.24† | 14.15 | 17.44† | 14.64 |
| pcNN50 (%) | 14.36 | 11.61 | 10.63 | 9.87 | 23.94*¥ | 18.79 | 6.58† | 8.73 | 6.58 | 9.40 | 6.77† | 8.74 |
| SDNN ($ms^2$) | 157.78 | 47.51 | 123.04* | 41.84 | 129.93* | 46.83 | 94.39† | 143.83 | 76.98† | 28.07 | 66.92† | 24.95 |
| RMSSD ($ms^2$) | 38.62 | 18.66 | 32.33 | 16.03 | 50.20 | 27.41 | 41.11 | 137.98 | 26.63 | 18.62 | 26.62 | 16.05 |
| SDANN ($ms^2$) | 142.99 | 54.74 | 104.65 | 57.11 | 92.20 | 38.02 | 127.15 | 602.03 | 61.80 | 23.36 | 46.95 | 20.65 |
| SDNNIDX ($ms^2$) | 67.90 | 22.28 | 64.27 | 21.07 | 75.36 | 28.12 | 52.82 | 110.72 | 41.27† | 16.29 | 41.99† | 15.49 |
| **Frequency domain** | | | | | | | | | | | | |
| Total power ($ms^2$) | 17151.33 | 15866.06 | 4163.78 | 2603.03 | 5815.03 | 4038.51 | 1112237.03 | 10510855.3 | 1757.52 | 1439.82 | 1814.43 | 1239.58 |
| VLF ($ms^2$) | 2629.77 | 1798.17 | 2298.41 | 1471.01 | 3270.90 | 2340.21 | 224403.18 | 2118248.78 | 1039.18 | 829.52 | 1127.80 | 746.23 |
| LF ($ms^2$) | 1248.20 | 745.96 | 1154.39 | 703.34 | 1437.10 | 968.93 | 16682.25 | 154756.07 | 381.71 | 385.13 | 388.20 | 349.87 |
| HF ($ms^2$) | 513.30 | 467.72 | 348.82 | 338.54 | 758.74 | 731.45 | 2653.18 | 23667.39 | 167.71 | 257.61 | 164.70 | 207.55 |
| LF (nu) | - | - | 79.02 | 8.29 | 65.95 | 12.13 | - | - | 72.56 | 18.84 | 70.96†§ | 21.16 |
| HF (nu) | - | - | 20.98 | 8.29 | 34.05¥ | 12.13 | - | - | 27.44† | 14.42 | 29.04†§ | 15.01 |
| LF/HF | - | - | 6.72 | 3.82 | 3.87†¥ | 2.80 | - | - | 5.39† | 3.98 | 5.39† | 4.47 |
| **Nonlinear domain** | | | | | | | | | | | | |
| Centroid (ms) | 794.10 | 102.19 | 737.47 | 94.59 | 936.65*¥ | 143.64 | 874.65† | 153.36 | 858.52† | 214.56 | 912.73 | 262.25 |
| SD1 (ms) | 26.62 | 12.80 | 22.22 | 10.86 | 35.03*¥ | 19.10 | 18.31† | 10.54 | 18.19 | 11.69 | 19.18† | 11.25 |
| SD2 (ms) | 215.71 | 66.15 | 167.14* | 53.51 | 175.26* | 64.19 | 110.13† | 33.81 | 104.97† | 34.65 | 94.66† | 41.21 |
| SD1/SD2 | 0.12 | 0.04 | 0.13 | 0.04 | 0.20*¥ | 0.07 | 0.17† | 0.09 | 0.17† | 0.10 | 0.21#§ | 0.12 |
| SD1 (nu) | 3.29 | 1.37 | 2.96 | 1.25 | 3.62¥ | 1.60 | 2.08† | 1.14 | 2.09† | 1.27 | 2.09† | 1.17 |
| SD2 (nu) | 27.00 | 6.79 | 22.53* | 6.03 | 18.60*¥ | 5.86 | 12.62† | 3.66 | 12.27† | 3.91 | 10.38† | 4.48 |
| Largest Lyapunov Exp. | 0.33 | 0.11 | 0.31 | 0.10 | 0.40*¥ | 0.15 | 0.25† | 0.15 | 0.23 | 0.10 | 0.24† | 0.10 |
| **Turbulence indexes** | | | | | | | | | | | | |
| Number of VPC's | 7.12 | 41.63 | - | - | - | - | 55.51† | 102.70 | - | - | - | - |
| Turbulence onset | -0.42 | 2.30 | - | - | - | - | 0.36† | 2.52 | - | - | - | - |
| Turbulence slope | 4.85 | 8.62 | - | - | - | - | 3.93 | 4.34 | - | - | - | - |
| Acceleration capacity | -8.47 | 2.14 | - | - | - | - | -5.43† | 2.09 | - | - | - | - |
| Deceleration capacity | 7.44 | 1.68 | - | - | - | - | 5.20† | 1.86 | - | - | - | - |

\* <0.05 versus 24h in Healthy group; ¥ between day and night in Healthy group; # <0.05 versus 24h in MI group; § between day and night in MI group; † <0.05 between Healthy and MI groups